\documentclass{article}
\usepackage{dsfont}
\usepackage{microtype}
\usepackage{graphicx}
\usepackage{subfigure}
\usepackage{booktabs} 
\usepackage{multirow}
\usepackage{placeins}
\usepackage{amsthm}
\usepackage{graphicx}
\usepackage{epstopdf}
\usepackage{threeparttable}
\usepackage{boldline}
\usepackage{cellspace}
\usepackage{siunitx} 
\usepackage{booktabs} 

\usepackage[ruled,linesnumbered]{algorithm2e}

\usepackage{hyperref}

\usepackage[accepted]{icml2023}

\usepackage{amsmath}
\usepackage{amssymb}
\usepackage{mathtools}
\usepackage{amsthm}
\usepackage[capitalize,noabbrev]{cleveref}

\theoremstyle{plain}
\newtheorem{theorem}{Theorem}[section]

\theoremstyle{definition}
\newtheorem{definition}[theorem]{Definition}

\theoremstyle{remark}

\usepackage[textsize=tiny]{todonotes}

\icmltitlerunning{Exposing the Fake: Effective Diffusion-Generated Images Detection}

\usepackage{amsmath,amsfonts,bm, bbm}
\usepackage{xparse}

\DeclareDocumentCommand\W{ g g }{%
        \IfNoValueTF {#1} {\mathbf{W}} {
            \IfNoValueTF {#2} {\mathbf{W}^{(#1)}}{\mathbf{W}^{(#1)}_{#2}}
        }
}

\DeclareDocumentCommand\bias{ g g }{%
        \IfNoValueTF {#1} {\mathbf{b}} {
            \IfNoValueTF {#2} {\mathbf{b}^{(#1)}}{\mathbf{b}^{(#1)}_{#2}}
        }
}

\DeclareDocumentCommand\betavar{ g g }{%
        \IfNoValueTF {#1} {\bm{\beta}} {
            \IfNoValueTF {#2} {{\bm{\beta}^{(#1)}}{}}{\bm{\beta}^{(#1)}_{#2}}
        }
}

\DeclareDocumentCommand\xivar{ g g }{%
        \IfNoValueTF {#1} {\bm{\xi}} {
            \IfNoValueTF {#2} {{\bm{\xi}^{(#1)}}{}}{\bm{\xi}^{(#1)}_{#2}}
        }
}

\DeclareDocumentCommand\xivarn{ g g }{%
        \IfNoValueTF {#1} {\bm{\xi^-}} {
            \IfNoValueTF {#2} {\bm{\xi^-}^{+(#1)}}{\bm{\xi^-}^{+(#1)}_{#2}}
        }
}

\DeclareDocumentCommand\xivarp{ g g }{%
        \IfNoValueTF {#1} {\bm{\xi^+}} {
            \IfNoValueTF {#2} {\bm{\xi^+}^{+(#1)}}{\bm{\xi^+}^{+(#1)}_{#2}}
        }
}

\DeclareDocumentCommand\nuvar{ g g }{%
        \IfNoValueTF {#1} {\bm{\nu}} {
            \IfNoValueTF {#2} {{\bm{\nu}^{(#1)}}{}}{\bm{\nu}^{(#1)}_{#2}{}}
        }
}

\DeclareDocumentCommand\hnuvar{ g g }{%
        \IfNoValueTF {#1} {\bm{\hat{\nu}}} {
            \IfNoValueTF {#2} {{\bm{\hat{\nu}}^{(#1)}}{}}{\bm{\hat{\nu}}^{(#1)}_{#2}{}}
        }
}

\DeclareDocumentCommand\muvar{ g g }{%
        \IfNoValueTF {#1} {\bm{\mu}} {
            \IfNoValueTF {#2} {{\bm{\mu}^{(#1)}}{}}{\bm{\mu}^{(#1)}_{#2}}
        }
}

\DeclareDocumentCommand\gammavar{ g g }{%
        \IfNoValueTF {#1} {\bm{\gamma}} {
            \IfNoValueTF {#2} {{\bm{\gamma}^{(#1)}}{}}{\bm{\gamma}^{(#1)}_{#2}}
        }
}

\DeclareDocumentCommand\lambdavar{ g g }{%
        \IfNoValueTF {#1} {\bm{\lambda}} {
            \IfNoValueTF {#2} {{\bm{\lambda}^{(#1)}}{}}{\bm{\lambda}^{(#1)}_{#2}}
        }
}

\DeclareDocumentCommand\tbetavar{ g g }{%
        \IfNoValueTF {#1} {{\bm{\tilde{\beta}}}} {
            \IfNoValueTF {#2} {{{\bm{\tilde{\beta}}}^{(#1)}}{}}{{{\bm{\tilde{\beta}}}^{(#1)}_{#2}}}
        }
}

\DeclareDocumentCommand\alphavar{ g g }{%
        \IfNoValueTF {#1} {\bm{\alpha}} {
            \IfNoValueTF {#2} {{\bm{\alpha}^{(#1)}}}{\bm{\alpha}^{(#1)}_{#2}}
        }
}

\DeclareDocumentCommand\D{ g g }{%
        \IfNoValueTF {#1} {\mathbf{D}} {
            \IfNoValueTF {#2} {\mathbf{D}^{(#1)}}{\mathbf{D}^{(#1)}_{#2}}
        }
}

\DeclareDocumentCommand\A{ g g }{%
        \IfNoValueTF {#1} {\mathbf{A}} {
            \IfNoValueTF {#2} {\mathbf{A}^{(#1)}}{\mathbf{A}^{(#1)}_{#2}}
        }
}

\DeclareDocumentCommand\AA{ g g }{
        \IfNoValueTF {#1} {\mathbf{\Omega}} {
            \IfNoValueTF {#2} {\mathbf{\Omega}(#1, #1)}{\mathbf{\Omega}(#1, #2)}
        }
}

\DeclareDocumentCommand\S{ g g }{%
        \IfNoValueTF {#1} {\mathbf{S}} {
            \IfNoValueTF {#2} {\mathbf{S}^{(#1)}}{\mathbf{S}^{(#1)}_{#2}}
        }
}

\DeclareDocumentCommand\K{ g g }{%
        \IfNoValueTF {#1} {\mathbf{K}} {
            \IfNoValueTF {#2} {\mathbf{K}^{(#1)}}{\mathbf{K}^{(#1)}_{#2}}
        }
}

\DeclareDocumentCommand\B{ g g }{%
        \IfNoValueTF {#1} {\mathbf{B}} {
            \IfNoValueTF {#2} {\mathbf{B}^{(#1)}}{\mathbf{B}^{(#1)}_{#2}}
        }
}

\DeclareDocumentCommand\lowerb{ g g }{%
        \IfNoValueTF {#1} {{\mathbf{\underline{b}}}} {
            \IfNoValueTF {#2} {{\mathbf{\underline{b}}}^{(#1)}}{{\mathbf{\underline{b}}}^{(#1)}_{#2}}
        }
}

\DeclareDocumentCommand\z{ g g }{%
        \IfNoValueTF {#1} {z} {
            \IfNoValueTF {#2} {z^{(#1)}}{z^{(#1)}_{#2}}
        }
}

\DeclareDocumentCommand\s{ g g }{%
        \IfNoValueTF {#1} {s} {
            \IfNoValueTF {#2} {s^{(#1)}}{s^{(#1)}_{#2}}
        }
}

\DeclareDocumentCommand\dom{ g g }{%
        \IfNoValueTF {#1} {\mathcal{S}} {
            \IfNoValueTF {#2} {\mathcal{S}_{#1}}{\mathcal{S}^{#1}_{#2}}
        }
}

\DeclareDocumentCommand\domlb{ g g }{%
        \IfNoValueTF {#1} {\mathsf{LB}} {
            \IfNoValueTF {#2} {\mathsf{LB}(\mathcal{#1})}{\mathsf{LB}(\mathcal{#1}_{#2})}
        }
}

\DeclareDocumentCommand\domub{ g g }{%
        \IfNoValueTF {#1} {\mathsf{UB}} {
            \IfNoValueTF {#2} {\mathsf{UB}(\mathcal{#1})}{\mathsf{UB}(\mathcal{#1}_{#2})}
        }
}

\DeclareDocumentCommand\uns{ g g }{%
        \IfNoValueTF {#1} {\tilde{s}} {
            \IfNoValueTF {#2} {\tilde{s}_{#1}}{s^{(#1)}_{#2}}
        }
}

\DeclareDocumentCommand\ub{ g g }{%
        \IfNoValueTF {#1} {u} {
            \IfNoValueTF {#2} {u^{(#1)}}{u^{(#1)}_{#2}}
        }
}

\DeclareDocumentCommand\lb{ g g }{%
        \IfNoValueTF {#1} {l} {
            \IfNoValueTF {#2} {l^{(#1)}}{l^{(#1)}_{#2}}
        }
}

\DeclareDocumentCommand\hz{ g g }{%
        \IfNoValueTF {#1} {\hat{z}} {
            \IfNoValueTF {#2} {\hat{z}^{(#1)}}{\hat{z}^{(#1)}_{#2}}
        }
}

\DeclareDocumentCommand\bu{ g g }{%
        \IfNoValueTF {#1} {\mathbf{u}} {
            \IfNoValueTF {#2} {\mathbf{u}^{(#1)}}{\mathbf{u}^{(#1)}_{#2}}
        }
}

\DeclareDocumentCommand\bl{ g g }{%
        \IfNoValueTF {#1} {\mathbf{l}} {
            \IfNoValueTF {#2} {\mathbf{l}^{(#1)}}{\mathbf{l}^{(#1)}_{#2}}
        }
}

\DeclareDocumentCommand\aaa{ g }{%
        \IfNoValueTF {#1} {\bm{a}} {
            {\bm{a}^{({#1})}}
        }
}

\DeclareDocumentCommand\haaa{ g }{%
        \IfNoValueTF {#1} {\bm{\hat{a}}} {
            {\bm{\hat{a}}^{({#1})}}
        }
}

\DeclareDocumentCommand\bbb{ g g }{%
        \IfNoValueTF {#1} {\mathbf{P}} {
            \IfNoValueTF {#2} {{\mathbf{P}_{#1}}}{{\mathbf{P}_{#1}^{({#2})}}}
        }
}

\DeclareDocumentCommand\hbbb{ g g }{%
        \IfNoValueTF {#1} {\mathbf{\hat{P}}} {
            \IfNoValueTF {#2} {{\mathbf{\hat{P}}_{#1}}}{{\mathbf{\hat{P}}_{#1}^{({#2})}}}
        }
}

\DeclareDocumentCommand\ccc{ g g }{%
        \IfNoValueTF {#1} {\mathbf{q}} {
            \IfNoValueTF {#2} {{\mathbf{q}_{#1}}}{{\mathbf{q}_{#1}^{(#2)}}{}}
        }
}

\DeclareDocumentCommand\constc{ g }{%
        \IfNoValueTF {#1} {c} {
            {c^{({#1})}}
        }
}

\DeclareDocumentCommand\setz{ g g }{%
        \IfNoValueTF {#1} {\mathcal{Z}} {
            \IfNoValueTF {#2} {\mathcal{Z}^{(#1)}}{\mathcal{Z}^{(#1)}_{#2}}
        }
}

\DeclareDocumentCommand\setzp{ g g }{%
        \IfNoValueTF {#1} {\mathcal{Z^+}} {
            \IfNoValueTF {#2} {\mathcal{Z}^{+(#1)}}{\mathcal{Z}^{+(#1)}_{#2}}
        }
}

\DeclareDocumentCommand\setzn{ g g }{%
        \IfNoValueTF {#1} {\mathcal{Z^-}} {
            \IfNoValueTF {#2} {\mathcal{Z}^{-(#1)}}{\mathcal{Z}^{-(#1)}_{#2}}
        }
}

\DeclareDocumentCommand\tsetz{ g g }{%
        \IfNoValueTF {#1} {\tilde{\mathcal{Z}}} {
            \IfNoValueTF {#2} {\tilde{\mathcal{Z}}^{(#1)}}{\tilde{\mathcal{Z}}^{(#1)}_{#2}}
        }
}

\DeclareDocumentCommand\tz{ g g }{%
        \IfNoValueTF {#1} {\tilde{z}} {
            \IfNoValueTF {#2} {\tilde{z}^{(#1)}}{\tilde{z}^{(#1)}_{#2}}
        }
}

\DeclareDocumentCommand\f{ g g }{%
        \IfNoValueTF {#1} {f} {
            \IfNoValueTF {#2} {f^{(#1)}}{f^{(#1)}_{#2}}
        }
}

\DeclareDocumentCommand\lf{ g g }{%
        \IfNoValueTF {#1} {\underline{f}} {
            \IfNoValueTF {#2} {\underline{f}^{(#1)}}{\underline{f}^{(#1)}_{#2}}
        }
}

\def\eqref#1{Eq.~(\ref{#1})}

\def\1{\bm{1}}

\DeclareMathAlphabet{\mathsfit}{\encodingdefault}{\sfdefault}{m}{sl}
\SetMathAlphabet{\mathsfit}{bold}{\encodingdefault}{\sfdefault}{bx}{n}

\begin{document}

\twocolumn[
\icmltitle{Exposing the Fake: Effective Diffusion-Generated Images Detection}




\begin{icmlauthorlist}
\icmlauthor{Ruipeng Ma}{uestc}
\icmlauthor{Jinhao Duan}{drexel}
\icmlauthor{Fei Kong}{uestc}
\icmlauthor{Xiaoshuang Shi}{uestc}
\icmlauthor{Kaidi Xu}{drexel}
\end{icmlauthorlist}

\icmlaffiliation{uestc}{Department of Computer Science and Engineering, University of Electronic Science and Technology of China}
\icmlaffiliation{drexel}{Drexel University}

\icmlcorrespondingauthor{Xiaoshuang Shi}{xsshi2013@gmail.com}

\icmlkeywords{Machine Learning, ICML}


\icmlkeywords{Machine Learning, ICML}

\vskip 0.3in
]



\printAffiliationsAndNotice{}  

\begin{abstract}
\noindent
Image synthesis has seen significant advancements with the advent of diffusion-based generative models like Denoising Diffusion Probabilistic Models (DDPM) and text-to-image diffusion models. Despite their efficacy, there is a dearth of research dedicated to detecting diffusion-generated images, which could pose potential security and privacy risks. This paper addresses this gap by proposing a novel detection method called \emph{Stepwise Error for Diffusion-generated Image Detection} (SeDID). Comprising statistical-based $\text{SeDID}_{\text{Stat}}$ and neural network-based $\text{SeDID}_{\text{NNs}}$, SeDID exploits the unique attributes of diffusion models, namely deterministic reverse and deterministic denoising computation errors. Our evaluations demonstrate SeDID's superior performance over existing methods when applied to diffusion models. Thus, our work makes a pivotal contribution to distinguishing diffusion model-generated images, marking a significant step in the domain of artificial intelligence security.
\end{abstract}

\section{Introduction}

Generative diffusion models have made significant strides in the field of image generation, demonstrating remarkable capabilities \cite{song2019generative, song2020score, ho2020denoising, rombach2022high, ramesh2022hierarchical, saharia2022photorealistic}, but have also raised privacy and abuse concerns~\cite{zhao2023unlearnable,kong2023efficient}.
Previous works~\cite{wang2023dire,ricker2022towards,corvi2022detection} have laid the groundwork for detecting diffusion-generated images, and some have successfully leveraged the deterministic reverse and denoising processes inherent to diffusion models.

However, while detection methods such as DIRE \cite{wang2023dire} indeed leverage some deterministic aspects, they may not fully exploit the entirety of these features. In particular, DIRE concentrates on the reconstruction at the initial $x_0$ timestep, which may overlook the valuable information encapsulated in the intermediate steps throughout the diffusion and reverse diffusion processes. In contrast, the proposed SeDID exploits these intermediate steps, which could potentially enhance the detection efficacy. Additionally, we reveal that the distribution of real images could potentially diverge from the distribution of images generated by diffusion models, given the inherently complex and diverse characteristics of natural images. This indicates that the real-image distribution might not align perfectly with the regular patterns learned by the diffusion process.

Given these observations, we distinctly formulate our research question as follows:

\emph{Can we discriminate between real and diffusion-generated images by harnessing the inherent distributional disparities between naturally occurring and diffusion-synthesized visuals?}

In our work, we address these issues by delving deeper into the deterministic reverse and denoising properties of diffusion models, proposing a novel and more encompassing detection approach. Our proposed method, the \emph{Stepwise Error for Diffusion-generated Image Detection} (SeDID), is designed to comprehensively utilize these unique diffusion properties to improve detection performance, thereby presenting a more generalized and robust solution for detecting diffusion-generated images.

Our approach draws inspiration from SecMI~\cite{duan2023diffusion}, a Membership Inference Attack (MIA) that differentiates training data and hold-out data on the assumption that the model overfits the training data. It's intuitive that a model better fits generation data than training samples is evident. Under this perception, we believe that the MIA-style method might be suitable for generation detection. Our method, which we have dubbed the \emph{Stepwise Error for Diffusion-generated Image Detection} (SeDID), utilizes the error between the reverse sample and the denoise sample at a specific timestep $T_{\textit{SE}}$.

\begin{figure*}[t]
\centering
\includegraphics[width=\textwidth]{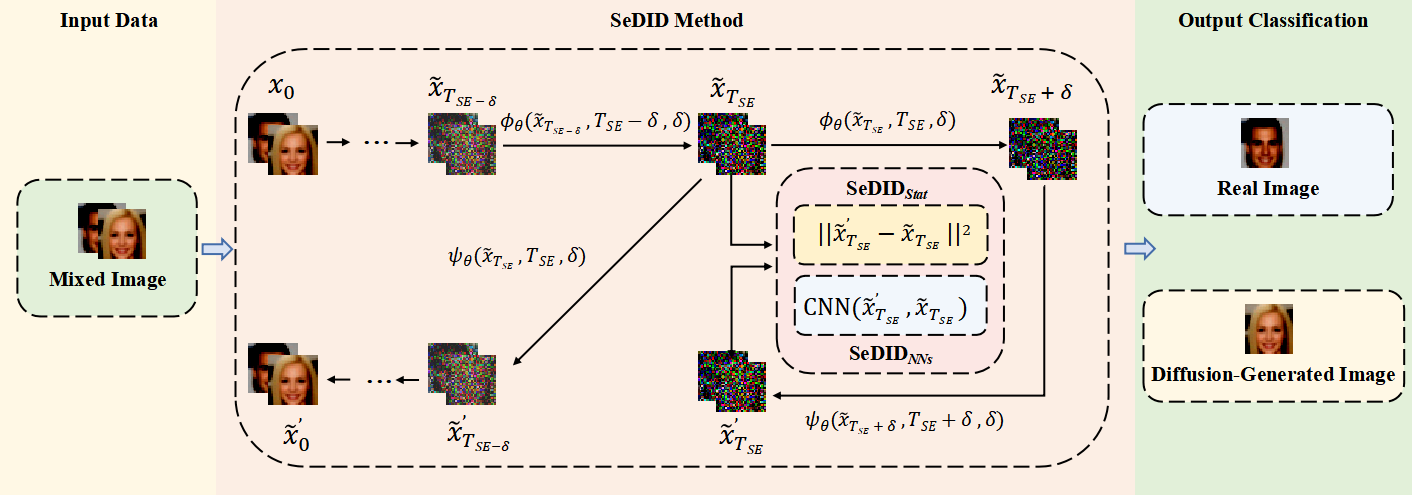}
\vspace{-6mm}
\caption{\textbf{The pipeline of our SeDID method.} Given mixed image data, it is processed through the SeDID method to compute the noise profile, a characterization of the noise patterns inherent to diffusion-generated images. Then, SeDID provides two branches: the Statistical-Based Synthetic Image Detection $\text{SeDID}_{\text{Stat}}$ and the Neural Network-Based Synthetic Image Detection $\text{SeDID}_{\text{NNs}}$. The $\text{SeDID}_{\text{Stat}}$ branch involves statistical analysis, error calculation, and model evaluation. The $\text{SeDID}_{\text{NNs}}$ branch employs a ResNet-18 model, which computes prediction errors, and updates weights via backpropagation. Both branches calculate the Area Under the Receiver Operating Characteristic Curve (AUC), the Accuracy (ACC), and the True Positive Rate at a given False Positive Rate (TPR@FPR), and classify images with \textit{real} or \textit{generated} output.}
\label{fig:pipeline}
\vspace{-3mm}
\end{figure*}

Our major contributions in this paper can be summarized as:

\begin{itemize}
  \setlength\itemsep{0.01em}
    \item We propose SeDID, a novel detection scheme for diffusion-generated images. SeDID uniquely exploits the distinct properties of diffusion models, particularly focusing on the errors between reverse and denoise samples at specific timesteps during the generation process.
    \item We adapt insights from membership inference attacks to emphasize the distributional disparities between real and generated data. This perspective enhances our understanding of diffusion models' security and privacy implications and underpins the design of SeDID.
    \item We present an extensive empirical evaluation of SeDID on three distinct datasets. The results demonstrate SeDID's superior performance in detecting diffusion-generated images, surpassing existing methodologies.
\end{itemize}

The remainder of this paper is organized as follows: Section 2 discusses the related work; Section 3 elaborates on our proposed methodology, SeDID; Section 4 presents the comprehensive evaluation of SeDID and discusses the results; finally, Section 5 concludes the paper and provides directions for future research.

\section{Related Works}

\subsection{Generative Diffusion Models}
Diffusion models, introduced by Sohl-Dickstein et al. \cite{sohl2015deep}, offer an approach distinct from Generative Adversarial Networks (GANs) \cite{goodfellow2020generative,yuan2020attribute,yuan2023dde}. These models gradually convert real data into noise and then learn to reverse this transformation. Ho et al. \cite{ho2020denoising} enhanced this process, leading to a better approximation of the real data distribution. Such improvement has significantly influenced our work, particularly our emphasis on the reverse and denoising steps. This field's versatility is demonstrated by Kong et al. \cite{kong2020diffwave}, who employed diffusion models in audio synthesis, inspiring our method's adaptability.

Diffusion models have been broadly employed for accelerating inference \cite{song2020denoising, salimans2022progressive, dockhorn2021score} and conditional generation \cite{dhariwal2021diffusion,ho2022classifier}. Several recent studies have addressed challenges such as improving inference speed and developing innovative methods \cite{xiao2021tackling,watson2022learning,rombach2022high,meng2021sdedit}.

\subsection{Diffusion-generated Image Detection}
Research on image detection originated with a focus on black-box \cite{shokri2017membership, salem2018ml, yeom2018privacy} and white-box attacks \cite{nasr2019comprehensive, rezaei2021difficulty}, both primarily developed for classification models \cite{sablayrolles2019white, song2021systematic, choquette2021label, hui2021practical, truex2019demystifying, salem2018ml, pyrgelis2017knock}. Black-box attacks assume limited knowledge about the model's internals, whereas white-box attacks presume complete model visibility. The research arena then widened to the detection of synthetic images, particularly those generated by diffusion models \cite{corvi2023detection}. This evolution incorporated the examination of forensic traces in diffusion-generated synthetic images and the performance evaluation of GAN-dedicated detectors when applied to these images, even in challenging contexts involving image compression and resizing in social networks.

DIRE \cite{wang2023dire} represents a preliminary exploration in detecting diffusion-generated data, utilizing the reconstruction error of images using Denoising Diffusion Implicit Models (DDIM) \cite{song2020denoising} for inversion and reconstruction. While this investigation is unfolding, other strides have been made within the broader field of diffusion models. Architectural advancements have been achieved with ADM \cite{dhariwal2021diffusion}, while PNDMs \cite{liu2022pseudo} have focused on accelerating the sampling speed. Furthermore, Stable Diffusion \cite{rombach2022high} v1 and v2 have delved into exploring downstream tasks.

\cite{duan2023diffusion} proposed the Step-wise Error Comparing Membership Inference (SecMI) approach for membership inference attack (MIA), leveraging the error comparison of the posterior estimation from the forward process. This concurrent work in the field inspired us in developing our current work, SeDID, which aims to detect diffusion-generated images effectively.

In summary, our SeDID method is a refined version of the concurrent work, SecMI, refocused specifically on detecting generated images, we adapt its technique to compute errors. This approach improves both the Area Under the Curve (AUC) and Accuracy (ACC) metrics in spotting diffusion-generated images, compared to methods solely relying on DDIM for image inversion and reconstruction.


\section{Methodology}

In this section, we detail our novel synthetic image detection method in diffusion models, namely SeDID, which builds upon the work of Duan~\cite{duan2023diffusion} and the Diffusion Denoising Probabilistic Models (DDPM)~\cite{ho2020denoising}. We start by defining key notations and outlining the fundamental principles of DDPM.

\subsection{Notations}

We use standard notations as defined by Ho et al. (2020)~\cite{ho2020denoising}. We denote the real data distribution as $q(x_0)$ and the latent variable model approximating $q(x_0)$ as $p_\theta(x_0)$. The ``noise-prediction'' model, $\epsilon_\theta$, is parameterized by weights $\theta$. 

The diffusion model comprises a $T$-step diffusion process $q(x_{t}|x_{t-1})$ and a denoising process $p_{\theta}(x_{t-1}|x_{t})$ for $1 \leq t \leq T$:
\begin{equation}
\begin{aligned}
q(x_{t}|x_{t-1}) = \mathcal{N}(x_{t}; \sqrt{1 - \beta_t}x_{t-1}, \beta_t I),\\
p_{\theta}(x_{t-1}|x_{t}) = \mathcal{N}(x_{t-1};\mu_{\theta}(x_t, t), \Sigma_{\theta}(x_t, t)),
\end{aligned}
\end{equation}
where $x_t$ refers to the diffusion result at timestep $t$, $\beta_t$ is the noise factor at timestep $t$, $I$ is the identity matrix, $\mu_{\theta}$ and $\Sigma_{\theta}$ are the mean and variance matrix of the denoising distribution respectively. The forward sampling at time step $t$ is:
\begin{equation}
q(x_t|x_0) = \mathcal{N}(x_t; \sqrt{\bar{\alpha}_t}x_0, (1 - \bar{\alpha}_t) I),
\end{equation}
\noindent where $\alpha_t = 1 - \beta_t$ and $\bar{\alpha}_t=\prod_{s=1}^t{\alpha_s}$.

\subsection{Definitions}

In the context of our work, we primarily focus on the deterministic denoising function $\psi_{\theta}$, and the deterministic reverse function $\phi_{\theta}$.We also introduce the notion of ($t\text{,}\delta$-\textit{error}), quantifying the posterior estimation error at timestep $t$ under stepsize $\delta$, and the Stepwise Error Calculation Time Step, $T_{\textit{SE}}$.

\begin{definition}[Deterministic denoising function $\psi_{\theta}$]\label{def:deterministic_denoising_function}
The deterministic denoising function $\psi_{\theta}(x, t)$, following the denoising process from DDIM~\cite{song2020denoising}, recovers the original data from the noised input $x$ at timestep $t$. It is defined as:
\begin{equation}\label{eq:deterministic_denoising_function}
\psi_{\theta}(x, t, \delta) = \sqrt{\bar{\alpha}_{t-\delta}}f_\theta(x, t) + \sqrt{1 - \bar{\alpha}_{t-\delta}}\epsilon_{\theta}(x, t),
\end{equation}
where $\epsilon_{\theta}(x, t)$ represents the stochastic noise, $\bar{\alpha}_t$ is the noise scale at timestep $t$ and $\psi_{\theta}(x, t, \delta)=x_{t-\delta}$. The definition of $f_\theta$ is~\eqref{eq:deterministic_forward_function}. We recover $x$ by applying the formula recurrently. Generally, we can use $\delta > 1$ to accelerate the denoising process.
\begin{equation}\label{eq:deterministic_forward_function}
f_\theta(x, t) = \frac{x - \sqrt{1 - \bar{\alpha}_t}\epsilon_\theta(x, t)}{\sqrt{\bar{\alpha}_t}}
\end{equation}
\end{definition}

\begin{definition}[Deterministic reverse function $\phi_{\theta}$]\label{def:deterministic_reverse_function}
The deterministic reverse function $\phi_{\theta}(x, t)$, following the reverse process from DDIM~\cite{song2020denoising}, is the reversed process of denoising process. Given a sample $x_t$, we can leverage $\phi_{\theta}(x, t)$ to obtain $x_{t+\delta}$:
\begin{equation}\label{eq:deterministic_reverse_function}
\phi_{\theta}(x, t, \delta) = \sqrt{\bar{\alpha}_{t+\delta}}f_\theta(x, t) + \sqrt{1 - \bar{\alpha}_{t+\delta}}\epsilon_{\theta}(x, t),
\end{equation}
where $\phi_{\theta}(x, t, \delta)=x_{t+\delta}$.
\end{definition}

The operations of $\phi_{\theta}$, $\psi_{\theta}$, and $f_{\theta}$ are applied during the diffusion process at specific timesteps determined by the Stepwise Error Calculation Time Step, $T_{\textit{SE}}$.

\begin{definition}[\label{def:t_error}$t\text{,}\delta$-\textit{error}] For a sample $x_0 \sim D$ and its deterministic reverse result $\tilde{x}_t=\phi_{\theta}(\phi_{\theta}(\dots\phi_{\theta}(x_0, 0, \delta), t-2\delta, \delta), t-\delta, \delta)$, we define ($t\text{,}\delta$-\textit{error}) as:
\begin{equation}\label{eq:t_error}
E_{t,\delta} = || \psi_{\theta} ( \phi_{\theta}(\tilde{x}_t, t, \delta), t, \delta) - \tilde{x}_{t} ||^2,
\end{equation}
where $\psi_{\theta}$ is the deterministic denoising function and $\phi_{\theta}$ is the deterministic reverse function.

In other word, we first use $\phi_{\theta}$ to obtain the $\tilde x_{t+\delta}$, and then, we use $\psi_{\theta}$ to get the reconstruction $\tilde x'_t$. $E_{t,\delta}$ is the difference beteen them. It is intuitive that the model is more familiar to the sample generated, so the sample with smaller $E_{t,\delta}$ will be more likely to be the sample generated.
\end{definition}
\begin{definition}[Stepwise Error Calculation Time Step, $T_{\textit{SE}}$ and stepsize, $\delta$]\label{def:stepwise_error_calculation_step}
$T_{\textit{SE}}$ is denoted to be the target timestep $t$ to compute ($t\text{,}\delta$-\textit{error}), and is called \textbf{Stepwise Error Calculation Timestep}. We abuse $\delta$ to represent the selected \textbf{stepsize}, since it is not ambiguous.
\end{definition}

\subsection{Detailed Design of the SeDID Method}

The SeDID method applies the concept of ($t\text{,}\delta$-\textit{error}) in combination with noise information during each diffusion process step, under the hypothesis that the process for real images differs from generated images. Experiments are conducted at various time steps $T_{\textit{SE}}$, representing different stages in the diffusion process.

At each experimental time step $T_{\textit{SE}}$, we calculate the ($t\text{,}\delta$-\textit{error}), $E_{T_{\textit{SE},\delta}}$ under stepsize $\delta$:
\begin{equation}
E_{T_{\textit{SE},\delta}} = || \psi_{\theta} ( \phi_{\theta}(\tilde{x}_{T_{\textit{SE}}}, T_{\textit{SE}}), T_{\textit{SE}},\delta) - \tilde{x}_{T_{\textit{SE}}} ||^2,
\end{equation}

\noindent where $x_{T_{\textit{SE}}}$ is the intermediate result from the diffusion process at the time step $T_{\textit{SE}}$, and $\tilde{x}_{T_{\textit{SE}}}$ is the corresponding image from the denoising process at the same time step.

\subsection{The Mechanics of the SeDID Method}

SeDID has two variants: Statistical-Based Synthetic Image Detection ($\text{SeDID}_{\text{Stat}}$) and the Neural Networks(NNs)-Based Synthetic Image Detection ($\text{SeDID}_{\text{NNs}}$).

\paragraph{Statistical-Based Synthetic Image Detection -- $\text{SeDID}_{\text{Stat}}$}

We apply ($t\text{,}\delta$-error) as the metrics to determine whether a sample is synthetic. If the error is smaller than a threshold $h$, we classify it to the synthetic sample. This can be described by:
\begin{equation}
    g(x) = \mathds{1}[E_{T_{\textit{SE}},\delta} < h]\label{g}
\end{equation}
The AUC, best ACC, and TPR@FPR are computed based on these noise profiles.

\paragraph{Neural Network-Based Synthetic Image Detection -- $\text{SeDID}_{\text{NNs}}$}

$\text{SeDID}_{\text{NNs}}$ extends the capabilities of $\text{SeDID}_{\text{Stat}}$ by incorporating a neural network with a ResNet-18 architecture. The network is fed with the intermediate outcomes of both the diffusion and reverse diffusion processes at timestep $T_{\textit{SE}}$. This allows the model to classify images as real or synthetic by using their respective noise profiles.

Our SeDID method leverages the strengths of statistical and machine learning approaches, accurately distinguishing real and synthetic images based on their noise profiles through a dual-phase strategy.

\section{Experiment}
In this section, we evaluate the performance of SeDID across various datasets and configurations.

\subsection{Datasets}

This study employed three publicly available datasets - CIFAR10~\cite{cifar10}, TinyImageNet~\cite{TinyIN}, and CelebA~\cite{CelebA} - each presenting distinct complexities due to their unique characteristics.

\textbf{CIFAR10:} This dataset contains 60,000 color images, evenly distributed across 10 classes. The set is divided into 50,000 training images and 10,000 validation images.

\textbf{TinyImageNet:} This subset of the ILSVRC-2012 classification dataset features 200 categories, each with 500 training images, 50 validation images, and 50 synthetic images. 

\textbf{CelebA:} This large-scale face attributes dataset comprises over 200,000 celebrity images, each annotated with 40 attribute labels. The varied size and complexity of the images in this dataset make it a challenging platform for generative models.

All images across all datasets in this work are preprocessed to a uniform size of 32 $\times$ 32 pixels.

\subsection{Experimental Settings}

\paragraph{Dataset Preparation}
\label{def:DataSet Preparation} All datasets underwent standard preprocessing operations, including normalization with a mean of 0.5 and standard deviation of 0.5 for each color channel, as well as potential random horizontal flipping for data augmentation. The CelebA dataset was further preprocessed with center cropping to 140 pixels as an additional augmentation step.

\paragraph{Baseline}
\label{def:Existing method}
We employ the existing Method~\cite{matsumoto2023membership} as our baseline. This approach uses the DDPM's training loss for membership inference by comparing generated Gaussian noise to predicted noise.

\paragraph{Metrics}
\label{def:Metrics}
Performance assessment of our SeDID hinged on three pivotal metrics: Accuracy (ACC), Area Under the ROC Curve (AUC), and True Positive Rate at a fixed False Positive Rate (TPR@FPR). ACC accounts for the fraction of correct predictions, AUC delineates model's class discriminative ability with higher scores signifying better performance, and TPR@FPR is an indicator of model's sensitivity at a given false-positive rate.

\paragraph{Experimental Setup and Implementation Details}
\label{def:Datasets and Diffusion Models}
In this work, we used diffusion models with the settings from \cite{ho2020denoising} and performed a diffusion process of $T=1000$ steps. The models were trained on the entire datasets with 1,200,000 steps for Tiny ImageNet and CelebA, and 800,000 steps for CIFAR10. We synthesized a balanced dataset of 50,000 diffusion-generated and 50,000 real images, selected from the training split of each dataset.

The $\text{SeDID}_{\text{NNs}}$ was trained on 10\% of this data, using a ResNet18 as the backbone. The training phase extended for 20 epochs at a learning rate of 0.001. To manage the learning dynamics, we employed Stochastic Gradient Descent (SGD) with a momentum of 0.9 and weight decay of 5e-4. The batch size was set to 128.

The computations were performed on an Intel(R) Xeon(R) Gold 6248R CPU @ 3.00GHz and an NVIDIA GeForce RTX 3090 GPU, ensuring the reproducibility of our experiments.

\subsection{Selection of Optimal $T_{\textit{SE}}$ and stepsize $\delta$}

In diffusion processes, the selection of optimal Stepwise Error Calculation Time Step, $T_{\textit{SE}}$, and stepsize $\delta$ is critical, as they directly affect the ($t\text{,}\delta$-error) - the difference between the expected and actual states of the synthetic image at time step $t$. These parameters are pivotal for optimizing the quality of synthetic image generation.

To determine the optimal $T_{\textit{SE}}$ and $\delta$, we experimented extensively on CelebA, Tiny-ImageNet, and CIFAR10 datasets. Our aim, by exploring various combinations, was to achieve the best possible AUC and ACC scores.

\begin{figure}[h]
\centering
\subfigure{\scalebox{0.27}{\includegraphics{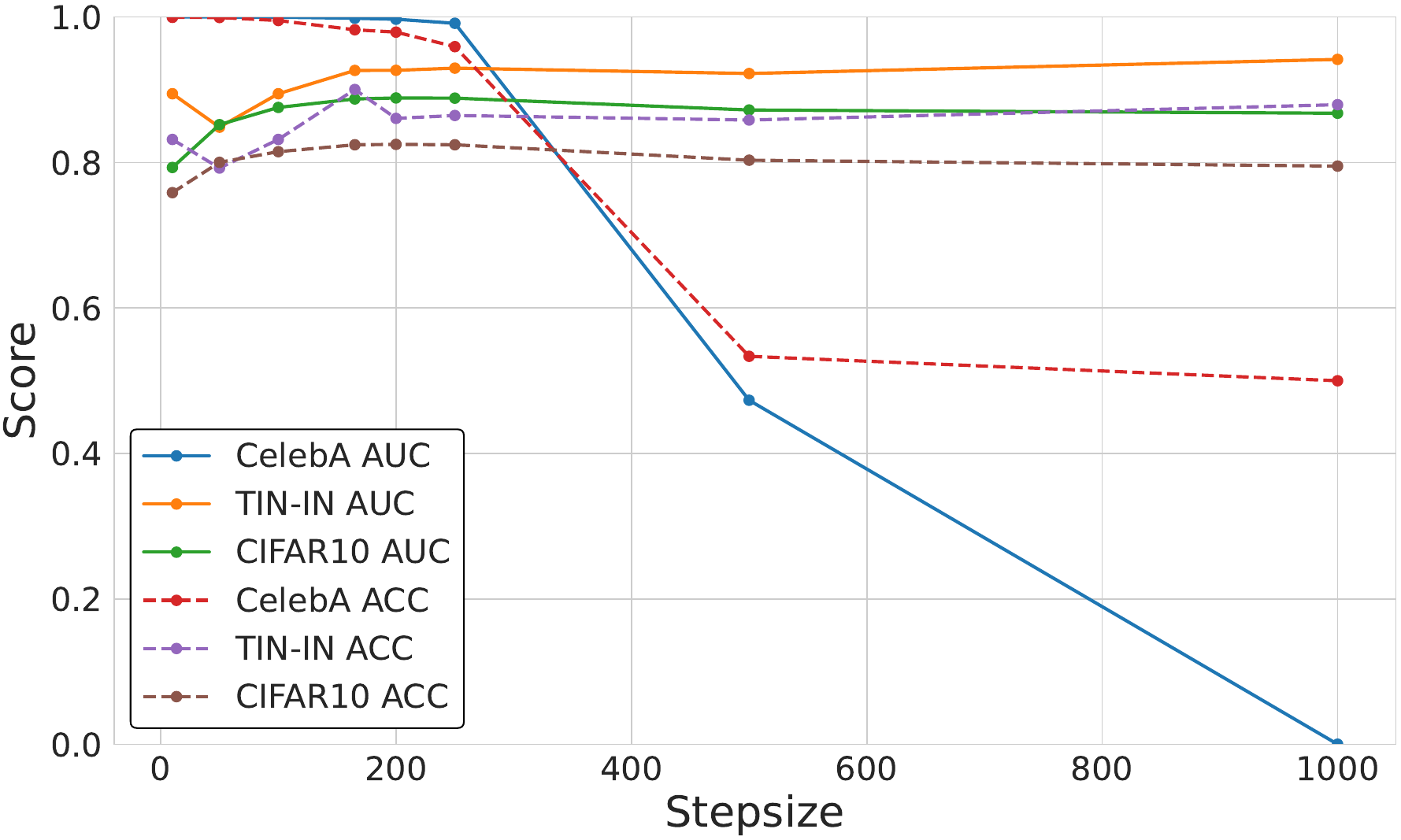}}}
\vspace{-3mm}
\caption{\textbf{Effect of different Stepsize on CelebA, TINY-IMAGENET, and CIFAR10 datasets.} The optimal Stepsize $\delta$ is highlighted with a triangle in each series.}
\label{fig:performance_Tse}
\end{figure}

Figure~\ref{fig:performance_Tse} depicts the performance of varying $\delta$ values across the datasets with best $T_{\textit{SE}}$. The optimal performance is achieved at $\delta=165$ for all datasets, thereby validating our choice.

\subsection{Experimental Results}

\begin{figure}[tbp]
\centering
\includegraphics[width=\columnwidth]{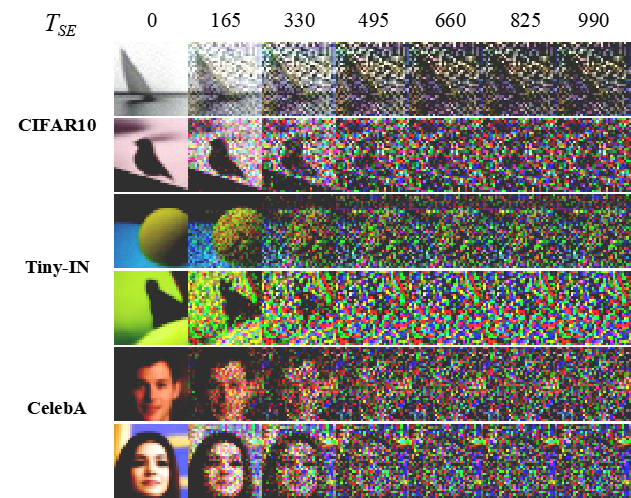}
\vspace{-4mm}
\caption{\textbf{Visual representation of the intermediate results of the SeDID method using the STAT strategy on the CIFAR10, Tiny-ImageNet, and CelebA datasets across various Stepwise Error Calculation Time Steps, $T_{\textit{SE}}$, within the range of [0, 1000] with the stepsize $\delta$ = 165.} This figure illustrates the diffusion progression under the SeDID method across different Stepwise Error Calculation Time Steps, $T_{\textit{SE}}$, at a fixed stepsize $\delta$.}
\label{fig:SeDID_visualization_datasets}
\vspace{-5mm}
\end{figure}

This subsection provides a comprehensive discussion of our experimental results conducted on three diverse datasets: CIFAR10, TinyImageNet, and CelebA. A visual representation of SeDID's performance across various Stepwise Error Calculation Time Steps, $T_{\textit{SE}}$, for the STAT strategy on these datasets is depicted in Figure \ref{fig:SeDID_visualization_datasets}.

As detailed in Table \ref{tab:performance_auc_acc}, SeDID consistently outperforms the existing method in detecting diffusion-generated images. This is demonstrated by its superior AUC and ACC values across all tested datasets. The results show the $T_{\textit{SE}}$ have significant impact to the performance. This observation underscores the necessity of selecting an optimal diffusion timestep to maximize detection performance and also highlights the potency of SeDID in effectively distinguishing diffusion-generated images.

\begin{table}[t]
    \caption{\textbf{Performance of GID-Diff across different timesteps.} The table shows the increasing AUC and ACC values with the growth of diffusion time $T$.}
    \label{tab:performance_auc_acc}
    \vspace{-2mm}
    \begin{center}
    \begin{small}
    \begin{sc}
    \resizebox{\columnwidth}{!}{
    \begin{tabular}{lccccccc}
        \toprule
        & & \multicolumn{2}{c}{CIFAR10} & \multicolumn{2}{c}{Tiny-ImageNet} & \multicolumn{2}{c}{CelebA} \\
        \cmidrule(lr){3-4}
        \cmidrule(lr){5-6}
        \cmidrule(lr){7-8}
        \multicolumn{1}{c}{$T$} & & AUC & ACC & AUC & ACC & AUC & ACC \\
        \midrule
        165 & & 0.4954 & 0.5089 & 0.1278 & 0.5000 & \bfseries 0.9985 & \bfseries 0.9843 \\
        330 & & 0.5415 & 0.5279 & 0.0606 & 0.5000 & 0.4213 & 0.5103 \\
        495 & & 0.5695 & 0.5579 & 0.5125 & 0.5240 & 0.2240 & 0.5001 \\
        660 & & 0.5667 & 0.5845 & 0.4971 & 0.6059 & 0.1866 & 0.5489 \\
        825 & & 0.8650 & 0.7992 & 0.9827 & 0.9615 & 0.0001 & 0.5000 \\
        990 & & \bfseries 0.8875 & \bfseries 0.8244 & \bfseries 0.9998  & \bfseries 0.9966 & 0.0000 & 0.5000 \\
        \bottomrule
    \end{tabular}
    }
    \end{sc}
    \end{small}
    \end{center}
    \vskip -0.1in
\end{table}

\begin{table*}[h]
    \caption{\textbf{SeDID variants and the existing method ~\cite{matsumoto2023membership} comparison on DDPM.} The table highlights the superior AUC and ACC of SeDID variants.}
    \label{tab:ddpm_3ds_auc_asr}
    \vspace{-1mm}
\begin{center}
\begin{small}
\begin{sc}
\resizebox{0.7\textwidth}{!}{
\begin{tabular}{lcccccccc}
    \toprule
     & \multicolumn{2}{c}{CIFAR10} & \multicolumn{2}{c}{Tiny-ImageNet} & \multicolumn{2}{c}{CelebA} & \multicolumn{2}{c}{Average}\\
     \cmidrule(lr){2-3}
     \cmidrule(lr){4-5}
     \cmidrule(lr){6-7}
     \cmidrule(lr){8-9}
     Method & AUC$\uparrow$  & ACC$\uparrow$  & AUC$\uparrow$  & ACC$\uparrow$  & AUC$\uparrow$  & ACC$\uparrow$  & AUC$\uparrow$  & ACC$\uparrow$  \\
     
     \midrule
     
     Existing method & 0.5293 &  0.5225 & 0.5303 & 0.5254 & 1.0000 & 1.0000 & 0.6865 & 0.6826 \\
     \midrule
     $\text{SeDID}_{\text{Stat}}$ & 0.8874 & \bfseries0.8244 & 0.9266 & 0.9004 & 0.9983 & 0.9825 & 0.9374 & 0.9024 \\
     $\text{SeDID}_{\text{NNs}}$ & \bfseries0.8903 & 0.8218 & \bfseries0.9999 & \bfseries0.9980 & \bfseries1.0000 & \bfseries1.0000 & \bfseries0.9634 & \bfseries 0.9399 \\
     \bottomrule
\end{tabular}
}
\end{sc}
\end{small}
\end{center}
    \vskip -0.1in
\end{table*}

\begin{figure}[ht]
\centering
\includegraphics[width=0.9\columnwidth]{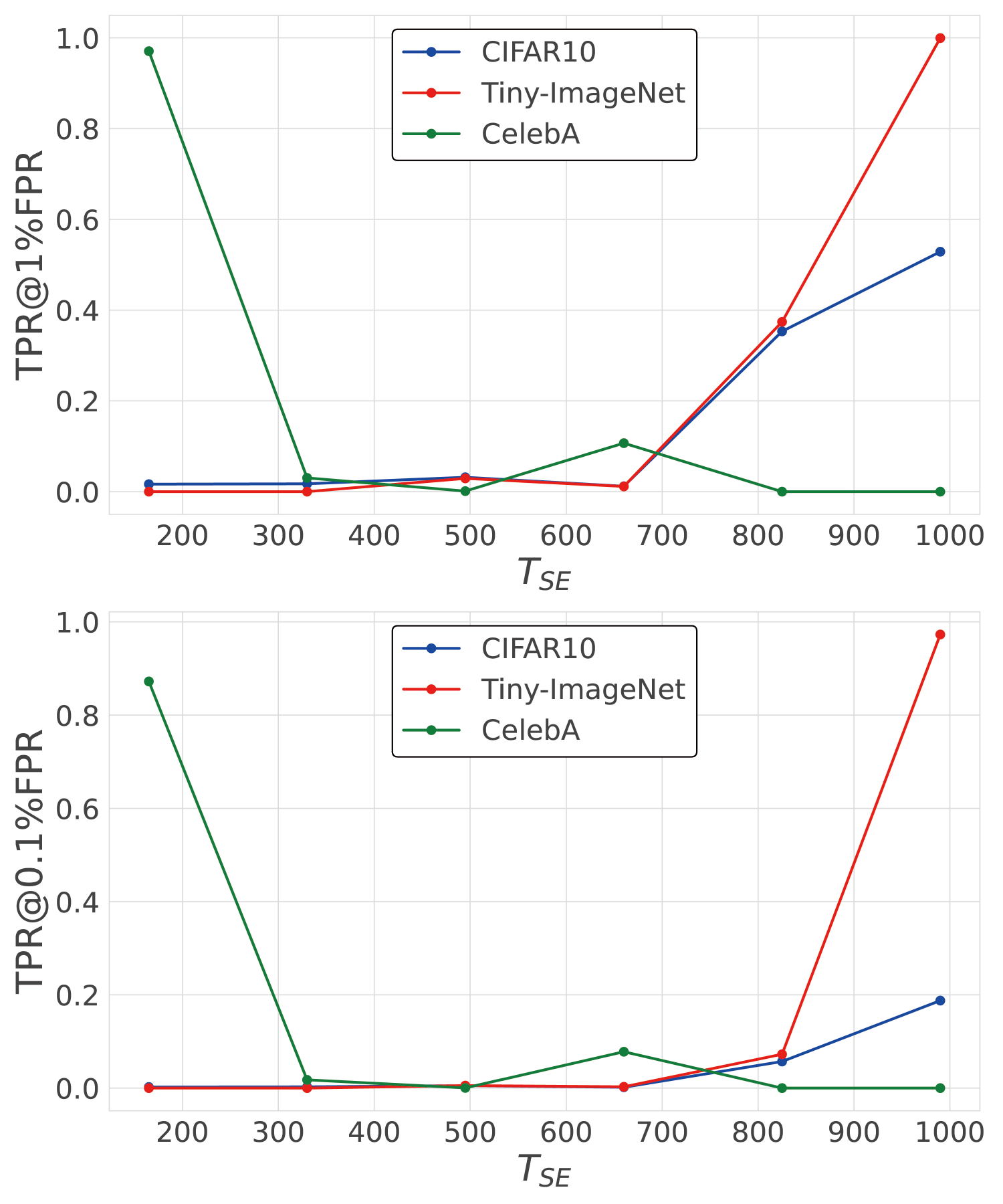}
\vspace{-4mm}
\caption{\textbf{Performance of the SeDID method at various timesteps.} This graph shows how SeDID's effectiveness varies with different time steps $T_{\textit{SE}}\in [0, 1000]$. Key observations include a smaller improvement rate in TPR@FPR(0.1\%) for CIFAR10 and Tiny-ImageNet as $T_{\textit{SE}}$ increases, and the highest performance for CelebA at small $T_{\textit{SE}}$.}
\label{fig:random_arg2}
\end{figure}

Additionally, we illustrate the diverse performance capabilities of our proposed SeDID method with respect to a range of different timesteps $T_{\textit{SE}}$ in Figure \ref{fig:random_arg2}. The results show as the value of $T_{\textit{SE}}$ progressively escalates, a distinct improvement in key performance metrics - TPR@FPR(1\%) and TPR@FPR(0.1\%) - is observed for the CIFAR10 and Tiny-ImageNet datasets. In stark contrast, these performance metrics exhibit a decreasing trend on the CelebA dataset with an increasing $T_{\textit{SE}}$. This intricate behavior not only underscores the inherent adaptability of the SeDID method, capable of fine-tuning $T_{\textit{SE}}$ to match the unique attributes of various datasets, but also hints at the potential dataset-dependence of the optimal $T_{\textit{SE}}$. This interpretation suggests that a meticulous, deliberate selection of $T_{\textit{SE}}$ might be a powerful strategy in significantly enhancing the performance of synthetic image detection tasks.

\subsection{Comparison to Baselines}

In the context of limited research specifically dedicated to the detection of images generated by diffusion models, we employ the existing method~\cite{matsumoto2023membership} as our benchmark. After evaluating our SeDID approach on the three datasets, the resulting AUC and ACC are presented in Table 2. It is evident that both $\text{SeDID}_{\text{Stat}}$ and $\text{SeDID}_{\text{NNs}}$ demonstrate superior performance over the existing method, particularly in terms of ACC with an average increase of over 30\% observed for both SeDID variants. This underscores the effectiveness of SeDID, particularly when employing neural networks in the inference strategy, as indicated by the notably high performance of SeDID$_{\text{NNs}}$.Furthermore, $\text{SeDID}_{\text{NNs}}$ displays an even more pronounced performance improvement, outperforming $\text{SeDID}_{\text{Stat}}$ by over 7\%. This demonstrates that training a neural network can significantly improve the performance in term of AUC and ACC.

\section{Conclusion and Future Work}

In this paper, we have presented SeDID, a novel method for detecting diffusion-generated images. SeDID leverages unique attributes of diffusion models, specifically deterministic reverse and deterministic denoising errors, providing a powerful tool for image detection. Extensive experiments on three different datasets demonstrate SeDID's superior performance compared to existing approaches. This work contributes to the field of diffusion-generated image detection, laying the groundwork for future research. Looking forward, we intend to:

\vspace{-2mm}
\begin{itemize}
  \setlength\itemsep{0.01em}
    \item Extend our approach to encompass other types of diffusion-based generation models, broadening its applicability.
    \item Investigate an automated mechanism for optimal timestep and $T_{\textit{SE}}$ selection, aiming to enhance the precision and effectiveness of detection.
    \item Explore the potential of a multi-step error computation approach in enhancing detection accuracy, leveraging our defined $t$-error concept.
    \item We plan to further investigate the influences on SeDID's performance across various datasets, specifically examining unusual trends in datasets like CelebA.
\end{itemize}
\vspace{-2mm}
Through these explorations, we aspire to develop more secure and reliable systems capable of effectively identifying and neutralizing potential threats from the misuse of generative diffusion models.

\bibliography{reference}


\clearpage
\appendix
\onecolumn
\section*{Appendix}


\section{Generalization to Latent Diffusion Model (LDM)}\label{appendix:generalize_various_diffusion}
For the Latent Diffusion Model (LDM), the calculation of $t$-\textit{error} is similar to DDPM, except the intermediate latent variables $v_t$ are in the latent space and the reverse process is conditioned by text embeddings.
Specifically, we denote by $V$ the Variational Autoencoders (VAEs) utilized to encode the original images into the latent space, i.e., $v_0 = V(x_0), x_0 \sim D $, and denote by $C$ the text condition.
The diffusion process and the denoising process can be derived as:
\begin{equation}
\begin{aligned}
q(v_{t}|v_{t-1}) = \mathcal{N}(v_{t}; \sqrt{1 - \beta_t}v_{t-1},  \beta_t \textbf{I}) \,\,\, \,\,\,\\
    p_{\theta}(v_{t-1}|v_{t}) = \mathcal{N}(v_{t-1};\mu_{\theta}(v_t, t, C),  \Sigma_{\theta}(v_t, t)).
\end{aligned}
\end{equation}
Then, the $t$-\textit{error} can be rewritten as:
\begin{equation}
    \tilde{\ell}_{t, v_0} = || \psi_{\theta} ( \phi_{\theta}(\tilde{v}_t, t, C), t, C) - \tilde{v}_{t} ||^2,
\end{equation}
where we reuse the symbols $\phi_{\theta}$ and $\psi_{\theta}$ as the deterministic reverse and sampling regarding $v_t$:
\begin{equation}
\begin{aligned}
    v_{t+1} & = \phi_{\theta}(v_t, t, C) \\
    &= \sqrt{\bar{\alpha}_{t+1}}f_\theta(v_t, t, C) + \sqrt{1 - \bar{\alpha}_{t+1}}\epsilon_{\theta}(v_t, t, C),
\end{aligned}
\end{equation}
\begin{equation}
\begin{aligned}
    v_{t-1} & = \psi_{\theta}(v_t, t, C) \\
    & = \sqrt{\bar{\alpha}_{t-1}}f_\theta(v_t, t, C) + \sqrt{1 - \bar{\alpha}_{t-1}}\epsilon_{\theta}(v_t, t, C),
\end{aligned}
\end{equation}
where
\begin{equation}
    f_\theta(v_t, t, C) = \frac{v_t - \sqrt{1 - \bar{\alpha}_t}\epsilon_\theta(v_t, t, C)}{\sqrt{\bar{\alpha}_t}}.
\end{equation}   

\section{Adopted Diffusion Model and Datasets}\label{appendix:adopted_model_datasets}
In our study, we employ CIFAE10, Tiny-ImageNet and CelebA on DDPM. The detailed settings are demonstrated in Table~\ref{tab:dataset_diffusion_maps}.

\begin{table*}[h]
    \caption{\textbf{Selected diffusion model and datasets.} This table outlines the model and datasets utilized in our study, detailing the image resolution and the count of real and generated images for each pair.}
    \label{tab:dataset_diffusion_maps}
    \vskip 0.15in
    \begin{center}
    \begin{small}
    \begin{sc}
    \begin{tabular}{lcccc}
        \toprule
         Model & Dataset & Resolution & \# Real Images & \# Generated Images\\
         \midrule
         \multirow{3}{*}{DDPM} & CIFAR-10 & 32 & 50,000 & 50,000 \\
         & Tiny-ImageNet & 32 & 50,000 & 50,000\\ 
         & CelebA & 32 & 50,000 & 50,000\\ 
         \bottomrule
    \end{tabular}
    \end{sc}
    \end{small}
    \end{center}
    \vskip -0.1in
\end{table*}

\section{Assessing the Effectiveness of Defensive Training in Diffusion Models}\label{appendix:failed_defense}

\begin{figure*}[htbp]
    \centering
    \includegraphics[width=0.85\columnwidth]{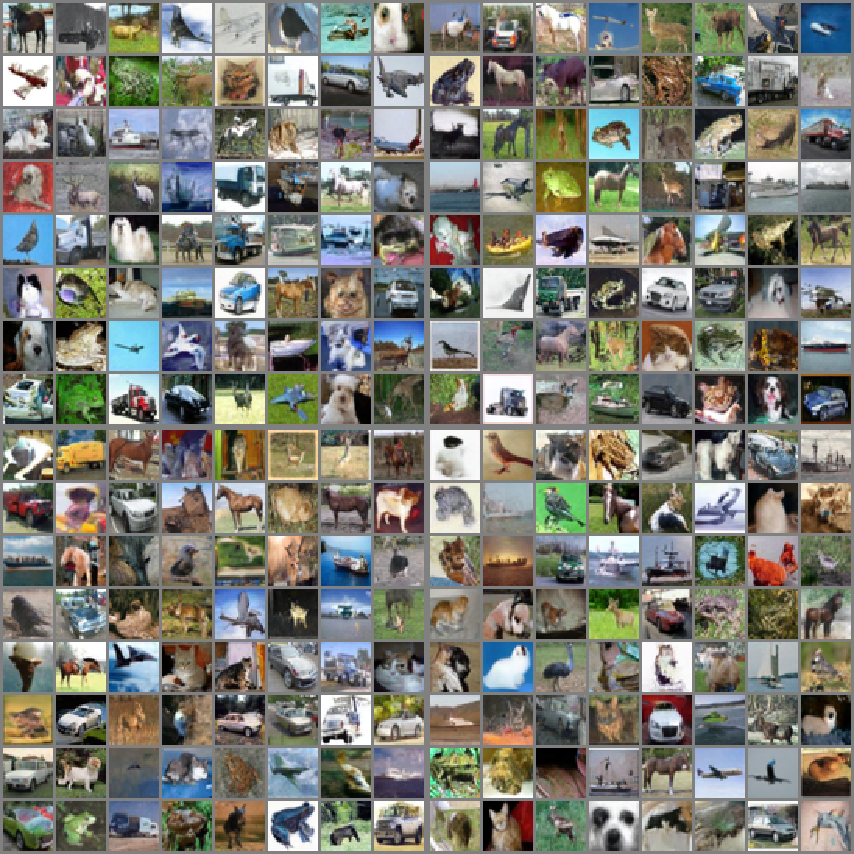}
    \caption{\textbf{Samples from DDPM on CIFAR-10 at step 800,000.}}
    \label{fig:cifar10}
\end{figure*}

\begin{figure*}[htbp]
    \centering
    \includegraphics[width=0.85\columnwidth]{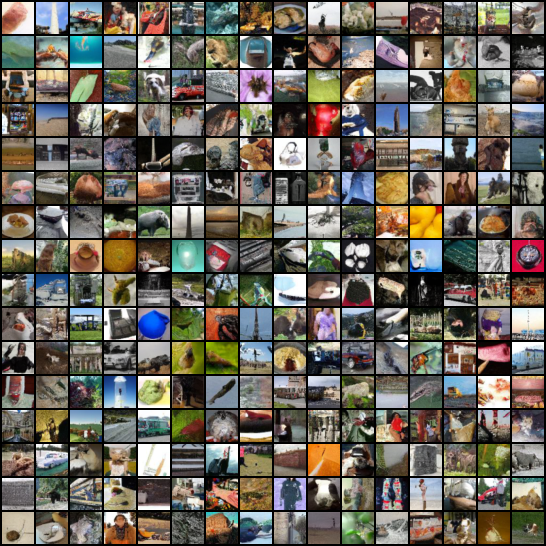}
    \caption{\textbf{Samples from DDPM on Tiny-ImageNet at step 1,200,000.}}
    \label{fig:TIN}
\end{figure*}

\begin{figure*}[t]
    \centering
    \includegraphics[width=0.85\columnwidth]{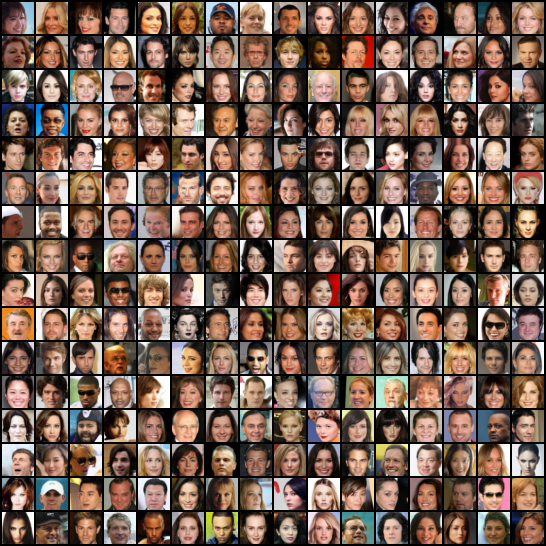}
    \caption{\textbf{Samples from DDPM on CelebA at step 1,200,000.}}
    \label{fig:CelebA}
\end{figure*}

We explore defensive training strategies for diffusion models and assess their effectiveness with the SeCID method. Specifically, we investigate aggressive regularization and data augmentation. The Figures~\ref{fig:cifar10}, ~\ref{fig:TIN}, and ~\ref{fig:CelebA} showcase samples from the resulting models.

Despite the defensive training efforts, the generated images are disappointingly vague and unrealistic, a consequence of the aggressive regularization and data augmentation. Moreover, the quality of reconstructed images from both the member and hold-out sets is significantly compromised, posing substantial challenges to the effectiveness of Membership Inference Attacks (MIA). The results underscore a fundamental trade-off between model performance and privacy protection in the context of diffusion models.

\end{document}